\documentclass[sigconf,authorversion]{acmart}

\copyrightyear{2024}
\acmYear{2024}
\setcopyright{rightsretained}
\acmConference[SIGCSE 2024]{Proceedings of the 55th ACM Technical Symposium on Computer Science Education V. 2}{March 20--23, 2024}{Portland, OR, USA}
\acmBooktitle{Proceedings of the 55th ACM Technical Symposium on Computer Science Education V. 2 (SIGCSE 2024), March 20--23, 2024, Portland, OR, USA}
\acmDOI{10.1145/3626253.3635608}
\acmISBN{979-8-4007-0424-6/24/03}
\AtBeginDocument{%
  \providecommand\BibTeX{{%
    \normalfont B\kern-0.5em{\scshape i\kern-0.25em b}\kern-0.8em\TeX}}}

\begin{document}

\title{Understanding the Role of Temperature in Diverse Question Generation by GPT-4}

\author{Arav Agarwal}
\orcid{0000-0001-9848-1663}
\affiliation{%
  \institution{Carnegie Mellon University}
  \city{Pittsburgh}
  \state{Pennsylvania}
  \country{USA}
}
\email{arava@andrew.cmu.edu}

\author{Karthik Mittal}
\orcid{0009-0005-5675-6987}
\affiliation{%
  \institution{Carnegie Mellon University}
  \city{Pittsburgh}
  \state{Pennsylvania}
  \country{USA}
}
\email{kkmittal@andrew.cmu.edu}

\author{Aidan Doyle}
\orcid{0009-0008-6260-7517}
\affiliation{%
  \institution{Carnegie Mellon University}
  \city{Pittsburgh}
  \state{Pennsylvania}
  \country{USA}
}
\email{adoyle@andrew.cmu.edu}

\author{Pragnya Sridhar}
\orcid{0000-0003-2160-288X}
\affiliation{%
  \institution{Carnegie Mellon University}
  \city{Pittsburgh}
  \state{Pennsylvania}
  \country{USA}
}
\email{pragnyas@andrew.cmu.edu}

\author{Zipiao Wan}
\orcid{0009-0002-5866-2376}
\affiliation{%
  \institution{Carnegie Mellon University}
  \city{Pittsburgh}
  \state{Pennsylvania}
  \country{USA}
}
\email{zwan@andrew.cmu.edu}

\author{Jacob Arthur Doughty}
\orcid{0009-0008-5430-7282}
\affiliation{%
  \institution{Carnegie Mellon University}
  \city{Pittsburgh}
  \state{Pennsylvania}
  \country{USA}
}
\email{jadought@andrew.cmu.edu}

\author{Jaromir Savelka}
\orcid{0000-0002-3674-5456}
\affiliation{%
  \institution{Carnegie Mellon University}
  \city{Pittsburgh}
  \state{Pennsylvania}
  \country{USA}
}
\email{jsavelka@andrew.cmu.edu}

\author{Majd Sakr}
\orcid{0000-0002-3739-298X}
\affiliation{%
  \institution{Carnegie Mellon University}
  \city{Pittsburgh}
  \state{Pennsylvania}
  \country{USA}
}
\email{msakr@institution.edu}

\renewcommand{\shortauthors}{Arav Agarwal et al.}

\begin{abstract}
  We conduct a preliminary study of the effect of GPT's temperature parameter on the diversity of GPT4-generated questions. We find that using higher temperature values leads to significantly higher diversity, with different temperatures exposing different types of similarity between generated sets of questions. We also demonstrate that diverse question generation is especially difficult for questions targeting lower levels of Bloom's Taxonomy. 
\end{abstract}

\maketitle

\section{Introduction}

Multiple Choice Questions (MCQs) are a common assessment tool in computing education. They allow for large-scale assessment, and can be graded automatically and, hence, efficiently.

However, developing effective MCQs remains an arduous task. Instructors need to devote significant time to craft question stems and answer choices in such a way that would allow for proper assessment of a student's understanding of the topic. 

There has been significant recent work into using large language models (LLMs) such as GPT-4 to generate educational materials, from student code explanations to auto-grader feedback \cite{10.1145/3545945.3569785, 10.1145/3593342.3593348}. One of the key parameters influencing the outcomes of the natural language generation of GPT models is \texttt{temperature}. The parameter generally determines the variety in generated text, with larger values generating more diverse text in other settings.  Despite its key role, there have not been many studies dedicated to exploring its effects on educational content generation. In this paper, we provide early results from our investigation into the effects of different \texttt{temperature} settings in an MCQ generation pipeline. Existing literature suggests that content generated by GPT-4 can be homogenous; temperature provides a way to improve the diversity of the generated content with minimal prompt engineering \cite{denny2023trust}.

\begin{table*}
\centering

\begin{tabular}{|r   |r   |r   |r |l |l|l|l|l|} \cline{1-4}\cline{6-9} 
Temperature & 1 Distinct Question & 2 Distinct Questions & 3 Distinct Questions   & & Temperature 
& N/A or I Don't Know& No&Yes
\\ \cline{1-4}\cline{6-9}   
0.2 & 34 & 46 & 22   & & 0.2 
& 1& 11&92
\\ \cline{1-4}\cline{6-9}  
1 & 17 & 17 & 68   & & 1 
&

1& 16&87\\ \cline{1-4}\cline{6-9}  
1.2 & 18 & 20 & 64   & & 1.2 & 
2& 15&87\\ \cline{1-4}\cline{6-9} 

\end{tabular}

\caption{Q1-distinct and Q2-complete vs. Temperature}
\vspace{-10pt}

\begin{tabular}{|r   |r|}\hline
 Bloom's Taxonomy Level&Percentage of Question Sets with Any Duplicates\\\hline
 Create&32\%\\\hline
  Analyze& 54\%\\ \hline 
  Apply& 55\%\\ \hline 
  Understand& 53\%\\ \hline 
  Remember& 67\%\\ \hline

\end{tabular}
\caption{Percentage of Question Sets with Duplicates vs. Taxonomy Level}
\vspace{-20pt}
\end{table*}

\section{Methodology}

We developed a system that generates MCQs, utilizing OpenAI's  GPT-4 \cite{10.1145/3636243.3636256}. Given a specific learning objective (LO), and high-level information about a course and a module, the system generates a single MCQ. 

In order to evaluate the effects of different temperature settings on generated questions, we gathered a dataset of 52 LOs spanning across multiple levels of Bloom's Taxonomy and common topics in introductory CS topics. Bloom's Taxonomy provides a hierarchical categorization of educational goals, and provides useful context to MCQ generation systems \cite{bloom1956taxonomy}. For each LO, we generated 3 MCQs of a randomly-chosen question type (either Fill-In-The-Blank, Code Tracing, Recall, Scenario or Identify Correct Output) across several values of \texttt{temperature} (0.2,1.0,1.2). These values were chosen to contain both the default value for GPT-4 (1.0), and the largest and the smallest reasonable values suggested by our initial experiments. We found that larger temperature values than 1.2 lead to generally unusable questions, while smaller temperature values than 0.2 simply rephrased the same question most of the time.
For each set of MCQs, we randomly sampled two instructors from a group of four and asked them to answer two questions: \emph{Q1-distinct}: How many of the generated MCQs were distinct? and \emph{Q2-complete}: Would it be possible to author an additional MCQ that is distinct from the generated ones but still aligned with the LO? \emph{Q1-distinct} focused on assessing the diversity of the set of generated MCQs, while \emph{Q2-complete} focused on assessing the completeness of the set. If an instructor is able to generate a distinct question, then there is potentially room for improvement, whereas if the instructor is unable to generate a new question then it is likely the LO is totally covered by the existing questions.

In total, we collected 313 annotations. Our overall inter-rater agreement was 0.30472 (Fleiss's $\kappa$), indicating we had fair agreement between instructors for both \emph{Q1-distinct} and \emph{Q2-complete} \cite{landis}. 

\section{Results}

We present the results of the preliminary study in Table 1 and Table 2. The left half of Table 1 counts the instructor responses to \emph{Q1-distinct} for each temperature setting, while the right half of Table 1 counts the instructor responses to \emph{Q2-complete}. When temperature is set to 0.2, we tend to generate more sets of questions where either two of the questions were too similar or all three questions were too similar. When temperature is set at 1.0 and 1.2, we tend to generate more sets of questions where all three questions are distinct. Comparatively, the instructor responses to \emph{Q2-complete} were the same at all temperature levels as, no matter the temperature setting, instructors generally answered that it was possible to create more distinct and aligned questions even with the generated question set.

We performed a chi-squared test and subsequent post-hoc analysis, and concluded that while the distributions of responses are significantly different comparing \texttt{temperature} 0.2 to the rest, there is not a significant difference between setting the \texttt{temperature} to 1.0 and 1.2. This suggests that, for the purpose of picking \texttt{temperature} values for quiz generation, one should generally strive to work with larger \texttt{temperature} values. Thus, in order to generate diverse questions using GPT-4, one should focus on temperature values between 1.0 and 1.2.

\section{Discussion}

Besides the above quantitative evaluation, we also performed a qualitative evaluation to discover trends, focusing on MCQs targeting different cognitive levels of Bloom's Taxonomy. 


We found that there appeared to be a relationship between the distinctness of MCQs and the targeted LO. If we look only at the level of Bloom's Taxonomy and look at the percentage of MCQ sets where \emph{Q1-distinct} identified multiple distinct questions (Table 2), we can see that, as we go to higher levels of Bloom's Taxonomy, instructors generally view the questions as having fewer duplicates. However, there is no clear relationship between completeness of generation (\emph{Q2-complete}) and Bloom's Taxonomy.
This suggests the issue with generation is not a problem with the LO itself, but with the generation process, and that GPT-4 might not be as capable in generating diverse MCQs at lower levels of Bloom's Taxonomy compared to higher levels of Bloom's Taxonomy. 

Thus, when looking at future studies in diverse question generation from LOs, one should focus on understanding LOs that are at lower levels of Bloom's Taxonomy, as they seem harder to generate automatically.

\section{Future Work}

Following this work, we would like to focus more heavily on various angles of the diverse question generation problem. Besides temperature, we would like to experiment with looking at the effect of other parameters like \texttt{frequency\_penalty} for diverse question generation, alongside differing prompting and LLM chaining strategies.

\bibliographystyle{ACM-Reference-Format}
\balance
\bibliography{sample-base}

\end{document}